\documentclass[conference]{IEEEtran}
\IEEEoverridecommandlockouts
\usepackage{cite}

\usepackage{tikz} 
\usepackage{listings}
\usepackage{enumitem}
\usepackage{graphicx}
\usepackage{multicol, latexsym}
\usepackage{blindtext}
\usepackage{subcaption}
\usepackage{caption}
\usepackage{longtable}

\usepackage{adjustbox}  

\usepackage{csquotes}
\usepackage{amsfonts}
\usepackage{amsmath}
\usepackage{amsthm}
\usepackage{amssymb}
\usepackage{algorithm}
\usepackage{tabularx}
\usepackage{capt-of,lipsum} 

\usepackage{listings}
\usepackage[draft]{hyperref}
\usepackage{comment}
\usepackage{diagbox}
\usepackage{pdflscape}
\usepackage{booktabs}
\usepackage{makecell}
\usepackage{balance}

\usepackage{array}

\usepackage{dirtytalk}
\usepackage{lipsum}
\usepackage{url}

\usepackage{mathtools, nccmath}

\usepackage{siunitx}
\DeclareSIUnit[number-unit-product = { }] \dBm{dBm}

\usepackage{algpseudocode}
\usepackage{algorithm}

\usepackage{listing-styles}

\algnewcommand\algorithmicforeach{\textbf{for each}}
\algdef{S}[FOR]{ForEach}[1]{\algorithmicforeach\ #1\ \algorithmicdo}
\usepackage[most]{tcolorbox}
\usepackage{multirow}

\definecolor{myblue}{rgb}{0.09,0.20,0.34}
\definecolor{mygreen}{rgb}{0,0.6,0}
\definecolor{mygray}{rgb}{0.98,0.98,0.98}
\definecolor{myorange}{rgb}{0.92,0.49,0.34}
\definecolor{mywhite}{rgb}{1.0,1.0,1.0}

\definecolor{NMR}{RGB}{255,255,86}
\definecolor{MRA}{RGB}{255,231,27}
\definecolor{MRD}{RGB}{178,178,178}
\definecolor{MRIP}{RGB}{188,172,0}
\definecolor{MRS}{RGB}{161,207,106}
\definecolor{NA}{RGB}{228,60,52}
\definecolor{AI}{RGB}{255,255,86}
\definecolor{RMR}{RGB}{162,4,21}
\definecolor{RMD}{RGB}{178,178,178}
\definecolor{RMA}{RGB}{255,231,27}
\definecolor{RMIP}{RGB}{188,172,0}
\definecolor{RMS}{RGB}{161,207,106}

\newsavebox{\mybox}

\usepackage{pifont}

\usepackage{censor}
\usepackage{wasysym}

\usepackage[scaled]{helvet}
\usepackage[T1]{fontenc}
\usepackage{helvet}

\usepackage{geometry}
\begin{document}



\title{Finding Pre-Injury Patterns in Triathletes from Lifestyle, Recovery and Load Dynamics Features}


\author{\IEEEauthorblockN{Leonardo Rossi, Bruno Rodrigues}
\IEEEauthorblockA{Embedded Sensing Group ESG\\
Institute of Computer Science in Vorarlberg ICV, University of St. Gallen HSG, Switzerland}
E-mail: leonardo.rossi@student.unisg.ch, bruno.rodrigues@unisg.ch
}


\maketitle

\begin{abstract}

Triathlon training, which involves high-volume swimming, cycling, and running, places athletes at substantial risk for overuse injuries due to repetitive physiological stress. Current injury prediction approaches primarily rely on training load metrics, often neglecting critical factors such as sleep quality, stress, and individual lifestyle patterns that significantly influence recovery and injury susceptibility. 

We introduce a novel synthetic data generation framework tailored explicitly for triathlon. This framework generates physiologically plausible athlete profiles, simulates individualized training programs that incorporate periodization and load-management principles, and integrates daily-life factors such as sleep quality, stress levels, and recovery states. We evaluated machine learning models (LASSO, Random Forest, and XGBoost) showing high predictive performance (AUC up to 0.86), identifying sleep disturbances, heart rate variability, and stress as critical early indicators of injury risk. This wearable-driven approach not only enhances injury prediction accuracy but also provides a practical solution to overcoming real-world data limitations, offering a pathway toward a holistic, context-aware athlete monitoring.

\end{abstract}


\section{Introduction}

Triathlon is a demanding multi-sport discipline that combines swimming, cycling, and running. Due to the high volume and intensity of training, athletes face a significant risk of overuse injuries, which result from cumulative training loads that exceed the body’s ability to recover and adapt~\cite{soligard2016much}. Multisensor wearable devices now enable continuous monitoring of such loads in daily life, particularly in unstructured, resource‑constrained environments. This injury risk is well-documented: in~\cite{andersen2013overuse}, 56\% of 174 participants in the Norseman Xtreme Triathlon reported overuse injuries during preparation for its 3.8\,km swim, 180\,km cycle, and 42\,km run.



Over the past decade, commercial wearables such as Garmin Forerunner and Polar Vantage have become de‑facto body-area sensor networks, streaming heart rate variability (HRV), resting heart rate, sleep staging, and training load to companion smartphones~\cite{mig2024impact}. These data streams offer valuable insights into fatigue and injury risk. However, extracting actionable patterns remains challenging~\cite{reis2024artificial}, significantly limiting their utility in injury prevention strategies. Recent work~\cite{jaiswal2024tinystressnet,kienstra2017triathlon} has demonstrated on-device processing for stress detection and related analytics, but injury forecasting in triathlon remains unexplored (cf. Section~\ref{relatedwork}).


Machine learning (ML) offers a promising approach to capture subtle physiological changes preceding injuries, improving over traditional threshold-based methods. ML has shown success in single-sport injury prediction, such as running~\cite{lovdal2021injury} and football~\cite{rossi2018effective,ayala2019preventive}. Triathlon introduces further complexity due to its multi-sport nature~\cite{hohl2024unveiling,kienstra2017triathlon}. Moreover, many existing studies focus narrowly on isolated metrics, overlooking broader factors like sleep, stress, and lifestyle, which strongly influence recovery and performance~\cite{halson2014monitoring}. In pervasive systems, such heterogeneous data must be fused under tight energy and bandwidth constraints, requiring generalisable and lightweight models.

Still, the broad use of ML in triathlon to predict overuse injuries faces significant barriers, primarily due to the limited availability of high-quality, systematically labeled injury datasets \cite{hohl2024unveiling}. Also, stringent ethical guidelines and privacy regulations further restrict access to sensitive athlete health data, posing substantial challenges for supervised ML methodologies \cite{lange2024generating}. We first introduced this research direction in a short paper \cite{rossi2025beyond}, with outlining the need for a holistic injury‑prediction and proposing an architecture. In this paper, we present the synthetic data generation framework archiecture, that fulfills this need of physiologically realistic athlete datasets. This framework simulates training loads, recovery dynamics, and injury occurrences with realistic physiological warning patterns, which allows the development and evaluation of ML-based injury prediction models without reliance on sensitive real-world data \cite{lange2024generating}. The primary contributions are the following:

\begin{itemize}
    \item The design of a synthetic athlete simulation framework grounded in established sports science principles of load management and periodization, incorporating realistic physiological responses and progressive injury patterns.
    \item An evaluation of multiple ML models for predicting injury events, emphasizing their capability to identify early physiological indicators preceding injury occurrences.
    \item We present how synthetic data generation can bridge methodological gaps imposed by privacy restrictions, providing a practical pathway toward deploying predictive models within real-world triathlon coaching and monitoring platforms.
\end{itemize}


\begin{table*}[!htbp]
    \centering
    \caption{Comparison of ML-Based Injury Prediction Approaches}
    \renewcommand{\arraystretch}{1.2}
    \resizebox{\textwidth}{!}{%
    \begin{tabular}{
        >{\centering\arraybackslash}p{2.8cm} 
        >{\centering\arraybackslash}p{2.4cm} 
        >{\centering\arraybackslash}p{2.6cm} 
        >{\centering\arraybackslash}p{2.2cm} 
        >{\centering\arraybackslash}p{1.8cm} 
        >{\centering\arraybackslash}p{1.8cm} 
        >{\centering\arraybackslash}p{1.8cm}}
        \toprule
        \textbf{Paper} & \textbf{Population} & \textbf{Data Modality} & \textbf{ML Technique} & \textbf{Recovery Features} & \textbf{Load Features} & \textbf{Synthetic Data} \\
        \midrule
        \cite{rossi2018effective} & Football & GPS, Accelerometer & DT + RFECV & \Circle & \CIRCLE & \Circle \\
        \cite{naglah2018athlete} & Football & Wearables, Surveys & SVM + KMeans & \CIRCLE & \CIRCLE & \Circle \\
        \cite{ayala2019preventive} & Football & Screening, Survey & ADTree + Ensemble & \CIRCLE & \LEFTcircle & \Circle \\
        \cite{lovdal2021injury} & Running & Wearables + Self-report & Bagged XGBoost & \LEFTcircle & \CIRCLE & \Circle \\
        \cite{rothschild2024predicting} & Endurance & Wearables + Apps & LASSO + Markov & \CIRCLE & \LEFTcircle & \Circle \\
        \cite{petrovsky2022tracking} & Multi-sport & Blood + Urine Biomarkers & RF + Logistic Reg. & \CIRCLE & \Circle & \Circle \\
        \cite{chen2024} & Multi-sport & Biomechanical, Environmental & Multi-clustering & \CIRCLE & \CIRCLE & \Circle \\
        \cite{kienstra2017triathlon} & Triathlon & Literature-based & None & \LEFTcircle & \LEFTcircle & \Circle \\
        \textbf{This work} & Triathlon & Simulated Wearable Data & LASSO, RF, XGBoost & \CIRCLE & \CIRCLE & \CIRCLE \\
        \bottomrule
    \end{tabular}%
    }
    \label{tab:injury_related_work}
\end{table*}

\section{Related Work}
\label{relatedwork}
Prior research in injury prediction span across several interconnected domains. We outline selected related work, summarizing their key methodological contributions and limitations in contrast to our work (cf. Table \ref{tab:injury_related_work}).

\subsection{Injury Prediction in Endurance Sports}
Kienstra et al. \cite{kienstra2017triathlon} explored injury risk in triathletes, emphasizing the dynamic interplay between internal and external load factors. They highlighted the potential utility of monitoring acute-to-chronic workload ratios, although empirical validation in triathlons remained incomplete. Halson \cite{halson2014monitoring} supported this perspective by highlighting internal load measures (e.g., heart rate, perceived exertion) as critical to understanding injury risk. Both emphasized the need for a holistic monitoring approach integrating multiple physiological metrics, an aspect central to our proposed framework.

Recent work has increasingly leveraged machine learning (ML) methods for injury prediction. Rothschild et al. \cite{rothschild2024predicting} applied LASSO regression with Markov unfolding techniques on wearable data from endurance athletes, achieving promising results for recovery prediction but facing variability challenges at the individual level. Similarly, Lövdal et al. \cite{lovdal2021injury} used XGBoost for runners and found superior performance with daily rather than weekly aggregated models, although their model was limited by the absence of objective recovery indicators.

Petrovsky et al. \cite{petrovsky2022tracking} demonstrated ML-based metabolic state classification from biomarkers using random forests and logistic regression, though their approach lacked direct relevance for daily injury prediction due to the absence of wearable sensor data.

\subsection{ML-Based Injury Prediction in Team Sports}
Several notable studies in football have applied ML to injury prediction with wearable technology. Rossi et al. \cite{rossi2018effective} achieved robust predictions of injury risk using GPS-derived load metrics and decision tree classifiers. Naglah et al. \cite{naglah2018athlete} developed a hierarchical SVM model with athlete-specific risk assessments based on internal and external load features. Ayala et al. \cite{ayala2019preventive} combined neuromuscular and psychological factors, employing ensemble learning methods to predict hamstring injuries effectively, though generalizability to endurance sports is limited.

\subsection{Synthetic Data Generation in Sports}
Recognizing the data scarcity challenge, synthetic data generation methods have recently gained attention. Hohl et al. \cite{hohl2024unveiling} compared multiple generative methods, finding TimeGAN superior in balancing fidelity, diversity, and predictive utility for recovery-related metrics. Lange et al. \cite{lange2024generating} demonstrated the effectiveness of CGAN for generating realistic stress-related wearable data, highlighting potential benefits and limitations regarding privacy preservation and diversity.

However, existing methods typically require initial real datasets and suffer from small sample sizes, raising concerns about overfitting and limited variability. Moreover, the application of synthetic data methods to triathlon-specific multimodal datasets remains unexplored.

Table \ref{tab:injury_related_work} summarizes previous approaches, highlighting key gaps that we address. Most existing studies focus on single-sport contexts (except \cite{petrovsky2022tracking, chen2024}), employ limited recovery metrics, or depend exclusively on real-world data, facing inherent privacy and scalability constraints. In contrast, our approach introduces a synthetic data generation framework explicitly tailored to triathlon contexts, incorporating comprehensive physiological and recovery metrics. This approach enables exploration of injury prediction in ways previously constrained by data scarcity.

We provide realistic triathlon-specific synthetic datasets, validated through extensive experimentation using ML algorithms (LASSO, Random Forest, and XGBoost) across both athlete-based and time-based data partitions. Furthermore, our work demonstrates the feasibility of synthetic data as a practical solution to overcome privacy restrictions, facilitating continuous model enhancement as real-world data becomes progressively available.

\section{Simulation Framework}
\label{sec:methodology}

We employ a comprehensive two-stage methodology comprising (A) synthetic athlete simulation and (B) supervised learning for injury prediction. 

\begin{table*}[htbp]
    \centering
    \caption{Parameter Ranges and Distributions to Build Athlete Profiles}
    \label{tab:athlete-parameters}
    \resizebox{\textwidth}{!}{
        \begin{tabular}{p{2.2cm}p{2.5cm}p{4cm}p{4cm}p{1.5cm}}
            \toprule
            \textbf{Category} & \textbf{Parameter} & \textbf{Range} & \textbf{Distribution} & \textbf{Source} \\
            \midrule
            \multirow{4}{*}{Demographic} & Gender & Binary & 60\% male, 40\% female & \cite{genderDistribution} \\
            & Age & 18--50 years & Normal ($\mu=33$, $\sigma=6$) & \cite{assis2024maximal}  \\
            & Height & Males: 164--192 cm & Normal ($\mu=178$, $\sigma=7$) & \cite{puccinelli2022performance}  \\
            & & Females: 153--177 cm & Normal ($\mu=165$, $\sigma=6$) & \cite{puccinelli2022performance} \\
            & Weight & Variable & Gender-specific, height-adjusted & \cite{puccinelli2022performance} \\
            \midrule
            \multirow{7}{*}{Physiological} & Genetic predisposition & 0.8--1.2 scale & Truncated normal ($\mu=1$, $\sigma=0.1$) & \cite{kellmann2018recovery} \\
            & HRV baseline & 40--95 ms & Age-adjusted, scaled by VO$_2$max, resting HR & \cite{plews2013training} \\
            & HRV range & ±15\% of baseline & Derived from baseline & \cite{plews2013training} \\
            & Resting HR & 38--60 bpm & Derived from VO$_2$max with lifestyle adjustments & \cite{bellenger2016monitoring} \\
            & Maximum HR & 165--205 bpm & Tanaka formula (208 - 0.7×age), random variation & \cite{tanaka2001age} \\
            & Lactate threshold HR & 160--190 bpm & ~87\% of max HR, adjusted for age, training & \cite{bentley2007incremental} \\
            & VO$_2$max & Males: 50--75 ml/kg/min & Normal ($\mu=62.5$, $\sigma=6.25$) & \cite{puccinelli2022performance} \\
            & & Females: 50--70 ml/kg/min & Normal ($\mu=60$, $\sigma=5$) & \cite{puccinelli2022performance} \\
            \midrule
            \multirow{3}{*}{Performance} & Functional threshold power & 2.5--5.5 W/kg & Gender-specific, adjusted for training experience & \cite{competitiveFTP} \\
            & Critical swim speed & 0.8--1.55 m/s & Converted to sec/100m (64--125 s/100m) & \cite{friel2010} \\
            & Running threshold pace & 3:00--5:30 min/km & VO$_2$max-correlated with adjustments & \cite{friel2010} \\
            \midrule
            \multirow{3}{*}{Training} & Experience level & 2--20 years & Right-skewed, age-constrained & \cite{competitivevolume} \\
            & Weekly training hours & 8--16 hours & Normal ($\mu=12$, $\sigma=2$), adjusted & \cite{competitivevolume} \\
            & Recovery rate & 0.5--1.3 scale & Function of age, VO$_2$max, lifestyle & \cite{kellmann2018recovery} \\
            \midrule
            \multirow{6}{*}{Lifestyle} & Sleep duration & 5--9 hours & Profile-dependent & \cite{halson2014sleep} \\
            & Sleep quality & 0.3--1.0 scale & Profile-dependent & \cite{castro2016sleep} \\
            & Nutrition quality & 0.4--1.0 scale & Profile-dependent & \cite{castro2016sleep} \\
            & Stress level & 0--0.9 scale & Profile-dependent & \cite{castro2016sleep} \\
            & Training adherence & 0.6--1.0 scale & Profile-dependent & \cite{castro2016sleep} \\
            & Smoking & 0--0.1 scale & 5\% occasional smokers & \cite{castro2016sleep} \\
            & Alcohol consumption & 0--0.6 scale & Profile-dependent & \cite{castro2016sleep} \\
            \bottomrule
        \end{tabular}
    }
\end{table*}

\begin{figure}[h]
    \centering
    \includegraphics[width=\columnwidth, trim={.7cm .9cm .7cm .9cm},clip]{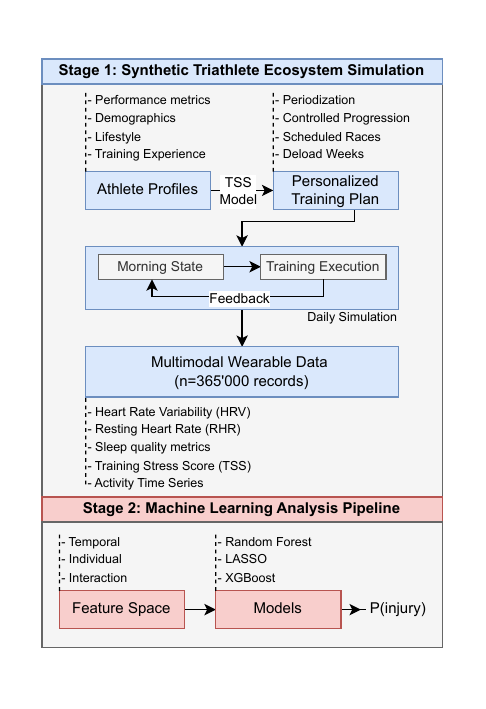}
    \caption{Methodology overview highlighting (A) synthetic data generation and (B) ML analysis stages.}
    \label{fig:methodology_overview}
\end{figure}

\subsection{Synthetic Triathlete Data Generator}

The synthetic data generator integrates four key components: athlete profile generation, personalized annual training plans, daily physiological simulations, and realistic injury pattern injections. Each step is grounded in robust empirical sports physiology literature, ensuring the simulated data accurately reflect real-world athlete behaviors and scenarios \cite{friel2010, kellmann2018recovery}. 

\begin{figure}[htbp]
    \centering
    \includegraphics[width=\columnwidth]{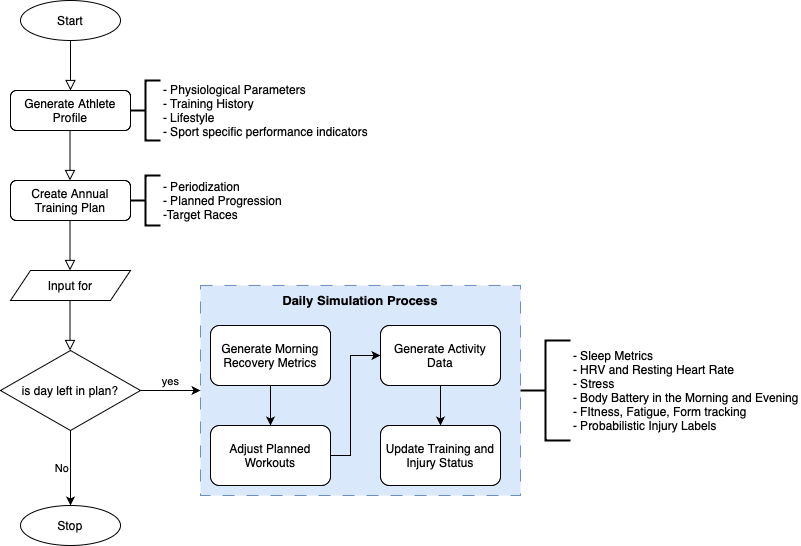}
    \caption{Data generation pipeline detailing morning assessment, workout execution, and recovery modeling.}
    \label{fig:data_generation_process}
\end{figure}

The data generation process includes (\textit{cf.} Figure \ref{fig:data_generation_process}):
\begin{itemize}
    \item Creating diverse athlete profiles with physiologically plausible characteristics
    \item Generating periodized annual training plans
    \item Simulating daily training dynamics including:
    \begin{itemize}
        \item Morning recovery metrics from wearable devices
        \item Realistic deviations from planned workouts based on physiological readiness
        \item Sport-specific wearable device activity data
        \item Training responses and potential injury events
    \end{itemize}
\end{itemize}

\subsubsection{Athlete Priors and Lifestyle Profiles}
We synthesize a diverse population of 1,000 competitive age-group triathletes, each assigned a profile with 24 parameters spanning demographic traits, physiological capacities, lifestyle characteristics, and sport-specific capabilities. Physiological values such as VO$_2$max, resting heart rate (HR), lactate threshold, and heart rate variability (HRV) are generated using empirically grounded distributions based on established literature \cite{assis2024maximal, bellenger2016monitoring}.

To reflect behavioral heterogeneity, we define six lifestyle archetypes including "Highly Disciplined Athlete" (30\%), "Balanced Competitor" (25\%), "Weekend Socialiser" (12\%), "Sleep-deprived Workaholic" (12\%), "Under-Recovered Athlete" (11\%) and "Health-Conscious Athlete" (10\%). Each profile identifies key lifestyle parameters such as sleep duration and quality, dietary habits, stress levels and alcohol consumption - all of which influence physiological metrics such as heart rate variability, recovery rate and performance markers \cite{castro2016sleep, young2018heart, ralevski2019heart}.

To maintain physiological realism and internal consistency, established principles of sport science were used to model interdependencies between parameters. This approach ensured that synthetic profiles represented plausible athlete characteristics rather than random combinations of values. For example, a multi-factorial approach was used to estimate lactate threshold heart rate (LTHR). 

\vspace{-1mm}
\begin{equation}
\small
\text{LTHR}_{\text{est}} = \max\left(160,\; \min\left(190,\; HR_{\text{max}} \cdot P_{\text{base}} \cdot M_{\text{final}} \cdot (1 + \epsilon)\right)\right)
\end{equation}
\vspace{-2mm}
where:
\begin{itemize}
    \item \( \text{LTHR}_{\text{est}} \) is the estimated lactate threshold heart rate.
    \item \( HR_{\text{max}} \) is the athlete's maximum heart rate.
    \item \( P_{\text{base}} = 0.87 \) is the base LTHR percentage for competitive triathletes.
    \item \( M_{\text{final}} \) is the final modifier, computed as:
\end{itemize}

\vspace{-2mm}
\begin{equation}
\small
M_{\text{final}} = M_{\text{age}} \cdot M_{\text{training}} \cdot M_{\text{VO2max}} \cdot M_{\text{HRR}} \cdot M_{\text{gender}}
\end{equation}
\vspace{-1mm}

Table \ref{tab:athlete-parameters} summarizes key input parameters, their statistical distributions, and sources from the literature. This include all physiological baselines, training experience, and recovery characteristics. Each athlete is assigned a unique identifier that serves as a primary key connecting all datasets.

The model begins with a baseline assumption that LTHR occurs at approximately 87\% of maximal heart rate in well-trained endurance athletes, consistent with the findings of Bentley et al. \cite{bentley2007incremental}, who demonstrated that lactate threshold typically manifests within 85-90\% of maximal heart rate in this population. This baseline is then modified by a series of evidence-based adjustments to account for key physiological determinants. For example, age-related modifiers are consistent with research by Tanaka and Seals \cite{tanaka2001age}, who found that while well-trained older athletes can maintain high lactate thresholds, age-related declines become more pronounced after the age of 50. Gender differences are accounted for with a slight upward adjustment for women (3\%), supported by Billaut and Bishop's \cite{billaut2009muscle} findings that women generally have higher fat oxidation rates and potentially superior lactate clearance at submaximal intensities. 

The model also incorporates VO$_2$max as a key determinant, following Coyle's \cite{coyle1995integration} research demonstrating that highly trained endurance athletes with superior maximal oxygen uptake can sustain work rates closer to 90\% of their VO$_2$max power. Finally, Heart Rate Reserve (HRR), the difference between maximal and resting heart rate, serves as an additional modifier, building on Karvonen's \cite{mj1957effects} foundational work establishing HRR as an effective predictor of cardiovascular efficiency.

\subsubsection{Periodized Season Planner}
Each athlete receives a 52-week training schedule adhering to well-established principles of periodization \cite{friel2010}. The schedule is divided into mesocycles (Base, Build, Peak, and Taper), with weekly \textit{Training Stress Score (TSS)} targets incrementally increasing, capped at 10\% to prevent overload \cite{kellmann2018recovery}. Weekly TSS distribution accounts for experience level and race calendar.

The training planner uses a rule-based scheduler that adjusts load intensity based on weekday preference, historical fatigue, and race proximity. Every fourth week within Build and Peak phases is programmed as a recovery week with a 30\% reduction in total TSS, aligning with known guidelines for avoiding overreaching and promoting supercompensation \cite{friel2010}.

\begin{figure}[htbp]
    \centering
    \includegraphics[width=\columnwidth]{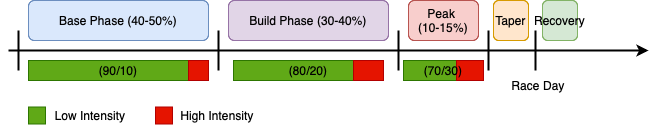}
    \caption{Training periodization across base, build, peak, and taper phases with controlled load increments.}
    \label{fig:training_periodization}
\end{figure}

Within each mesocycle, the sport-specific TSS breakdown—swimming (20\%), cycling (50\%), running (25\%), strength (5\%)—is adapted according to athlete preference and performance goals.

\subsubsection{Daily Physiology Simulation}
For each day, the simulator updates athlete state variables and produces wearable-device-compliant output. This includes heart rate, power, pace, sleep metrics, and stress indicators.

\begin{itemize}
    \item \textbf{Morning recovery}: HRV, resting HR, and sleep quality are sampled from conditional distributions based on prior-day training load, cumulative fatigue, and stress.
    \item \textbf{Workout execution:}: the completion probability of each workout is determined using a logistic function of fatigue and HRV trend. Executed workouts generate 15-second resolution time-series for heart rate, pace, and power output \cite{storniolo2020heart}.
    \item \textbf{Post-Workout metrics}: fitness, fatigue, and form are computed using exponentially weighted moving averages of adjusted TSS \cite{murray2017calculating}.
\end{itemize}

\subsubsection{Injury Pattern Injection}
Injury simulation incorporates both progressive and acute patterns, doing the assignment of injury labels to each day. The label is binary, where 1 indicates an injury on a given day and 0 indicates no injury. This component represents one of the most significant challenges in the simulation process, as positive injury labels must be preceded by realistic physiological patterns rather than occurring randomly. 

\begin{figure}[htbp]
    \centering
    \includegraphics[width=1\columnwidth]{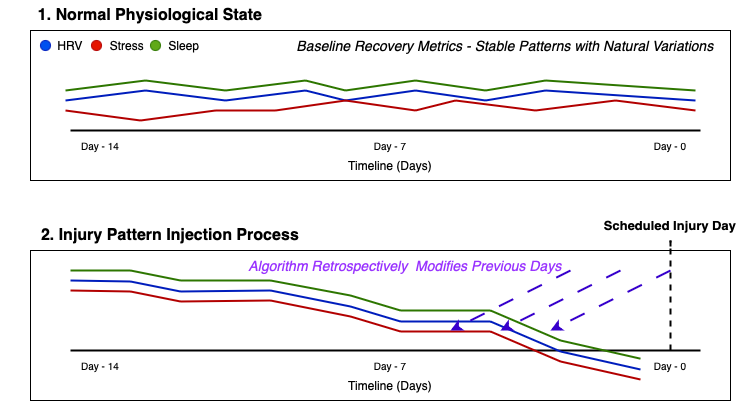}
    \caption{Injury pattern injection mechanism simulating degradation in recovery metrics prior to labeled injury events.}
    \label{fig:injury_pattern_injection}
\end{figure}

To address this challenge, the framework evolved from a probabilistic model that assigned injury probabilities based on previous recovery and performance data, to a system of scheduled injuries, that once reached lead to retrospective pattern injection (\textit{cf.} Figure \ref{fig:injury_pattern_injection}) Each athlete receives 4--6 pattern-based injuries annually. These are preceded by physiological degradations across HRV, resting HR, and sleep quality within a 7–14 day window \cite{plews2013training}. To prevent overfitting, 2--3 false-alarm periods are also introduced, containing warning signals but no actual injury.

Each injected injury triggers a post-injury recovery window during which training intensity is automatically reduced and injury labels are active. During this period, wearable indicators progressively normalize.

These patterns are based on established physiological warning signs that typically precede athletic injuries \cite{plews2012heart, plews2013training, buchheit2014monitoring, bellenger2016monitoring, milewski2014chronic, fullagar2015sleep, ivarsson2017psychosocial}. Recent research has shown that measurable changes in several biomarkers often occur days or weeks before an injury becomes symptomatic.

The primary physiological metrics that exhibit pre-injury patterns include:

\begin{itemize}
    \item \textbf{Heart Rate Variability (HRV)}: typically showing a progressive decline \cite{plews2013training, plews2012heart}
    \item \textbf{Resting Heart Rate (RHR)}: often demonstrating a gradual increase \cite{buchheit2014monitoring, bellenger2016monitoring}
    \item \textbf{Sleep quality metrics}: including disruptions to REM and deep sleep phases \cite{milewski2014chronic, fullagar2015sleep}
    \item \textbf{Stress markers}: generally increasing as the body struggles to adapt \cite{ivarsson2017psychosocial}
\end{itemize}

\subsection{Data and Feature Engineering}

The output from the data generation framework in the previous section consists of three interrelated CSV files representing a comprehensive dataset of 1000 athletes over the full calendar year of 2024:
\begin{itemize}
    \item \textbf{athletes.csv}: foundational athlete profiles with all parameters specified in Table \ref{tab:athlete-parameters}. Each athlete is assigned a unique identifier that serves as a primary key connecting all datasets. 
    \item \textbf{daily\_data.csv}: records day-by-day physiological signals captured by simulated wearable devices. Each record includes metrics such as HRV, resting heart rate, sleep quality, stress levels, recovery indicators and a binary injury label. 
    \item \textbf{activity\_data.csv}: documents every completed training activity with detailed metrics including duration, training load, and sport type as well as sport-specific metrics. 
\end{itemize}

\subsubsection{Injury Label Engineering}

The raw injury data, represented as a binary indicator (1 = injured, 0 = not injured), has several limitations for predictive modeling. In particular, this simplistic representation combines both the initial onset of injuries and the subsequent recovery period into a single category, obscuring the distinctive physiological patterns that precede injuries. To address this limitation, we transformed raw injury indicators into more nuanced and meaningful representations seen in Table \ref{tab:data_generation_singlecolumn}.

\begin{table}[t]
\centering
\caption{Overview of Injury Labeling}
\label{tab:data_generation_singlecolumn}
\renewcommand{\arraystretch}{1.15}
\resizebox{\columnwidth}{!}{%
\begin{tabular}{@{}p{2.4cm}p{1.2cm}p{4.8cm}@{}}
\toprule
\textbf{Name} & \textbf{Type / Range} & \textbf{Purpose} \\ \midrule
\texttt{injury\_onset} & Boolean & Marks the first symptomatic day of each injury episode. \\
\texttt{pre\_injury} & Boolean & Flags the seven days preceding an onset to enable early warning. \\
\texttt{recovery} & Boolean & Identifies days after onset while the athlete is still injured; masks post‑event data to avoid leakage. \\
\texttt{injury\_state} & Categorical $\{0,1,2,3\}$ & Compact multi‑class code (0 = healthy, 1 = pre‑injury, 2 = onset, 3 = recovery). \\
\texttt{will\_get\_injured} & Boolean & Forward‑looking target: injury expected within the next 7 days. \\
\texttt{time\_to\_injury} & Integer [0,28] & Days until the next injury episode; capped at 28 d for stability. \\ \bottomrule
\end{tabular}}
\end{table}

First, it distinguishes between injury onset and recovery, preventing the model from learning patterns associated with the recovery process rather than prediction. Second, the forward-looking "will\_get\_injured" target directly aligns with the practical goal of early intervention. Third, by creating a multi-class representation, this injury labeling enables more nuanced analysis of the model's predictive capabilities across different phases of the injury cycle.


\subsubsection{Feature Set}
Over 100 features are derived from daily and rolling statistics across physiological and training metrics:
\begin{itemize}
    \item Rolling min, max, mean, and slope (3, 7, 14-day windows) for HRV, HR, sleep, TSS.
    \item Recovery ratios: current HRV divided by athlete baseline.
    \item Risk markers: Acute-Chronic Workload Ratio, load spikes, monotony, strain \cite{gabbett2016training}.
    \item Interaction terms: Sleep $\times$ HRV, Stress $\times$ Load.
\end{itemize}

To capture complex interactions between physiological systems, compound features were developed. The features of the sleep-HRV compound identified days when both sleep quality and autonomic nervous system regulation were compromised simultaneously. Load-recovery interactions highlighted imbalances between training stress and recovery capacity. Stress-recovery ratios quantified the relationship between psychological stress levels and physiological recovery metrics, capturing how external stressors might amplify training-related fatigue.

\subsubsection{Models and Evaluation}
We evaluate three established classifiers: Random Forest \cite{breiman2001random}, LASSO Logistic Regression \cite{tibshirani1996regression}, and XGBoost \cite{chen2015xgboost}. All models incorporate class weighting to account for the class imbalance.

Evaluation uses two complementary validation schemes:
\begin{itemize}
    \item \textbf{Athlete-based split:} 80\% athletes for training, 20\% unseen athletes for testing.
    \item \textbf{Time-based split:} Train on Jan--Oct, test on Nov--Dec, across all athletes.
\end{itemize}

All models were implemented in Python 3.8 using scikit-learn (1.6.1) \cite{scikit-learn} for the LASSO and Random Forest algorithms, and XGBoost (3.0.0) \cite{Chen:2016:XST:2939672.2939785} for the gradient boosting implementation. 

The LASSO model employed logistic regression with L1 regularization using the 'liblinear' solver, configured with a convergence tolerance of 0.001 and maximum 1000 iterations. For the Random Forest classifier, an ensemble of 200 trees was constructed with a maximum depth of 8 and minimum samples per leaf set to 7 to balance complexity and generalization. The XGBoost implementation utilized 400 estimators with shallow trees (maximum depth of 2), a conservative learning rate of 0.03, and subsampling parameters (0.8 for observations, 0.7 for features) to mitigate overfitting. All models incorporated class weighting to address the substantial imbalance between injury and non-injury observations, with weights inversely proportional to class frequencies.

\section{Evaluation}
\label{sec:evaluation}

Our evaluation focuses on answering three major points:
\begin{itemize}
    \item \textbf{EQ1}: \emph{Physiological plausibility} – Do the synthetic signals stay inside accepted athlete ranges?
    \item \textbf{EQ2}: \emph{Pre‑injury realism} – Do synthetic trajectories reproduce the week‑scale drifts that precede real injuries?
    \item \textbf{EQ3}: \emph{Predictive benefit} – Does adding synthetic data improve early‑injury ML prediction when real data are scarce?
\end{itemize}

\subsection{Physiological Plausibility (EQ1)}


\begin{figure}[h!]
\centering
\includegraphics[width=1\columnwidth]{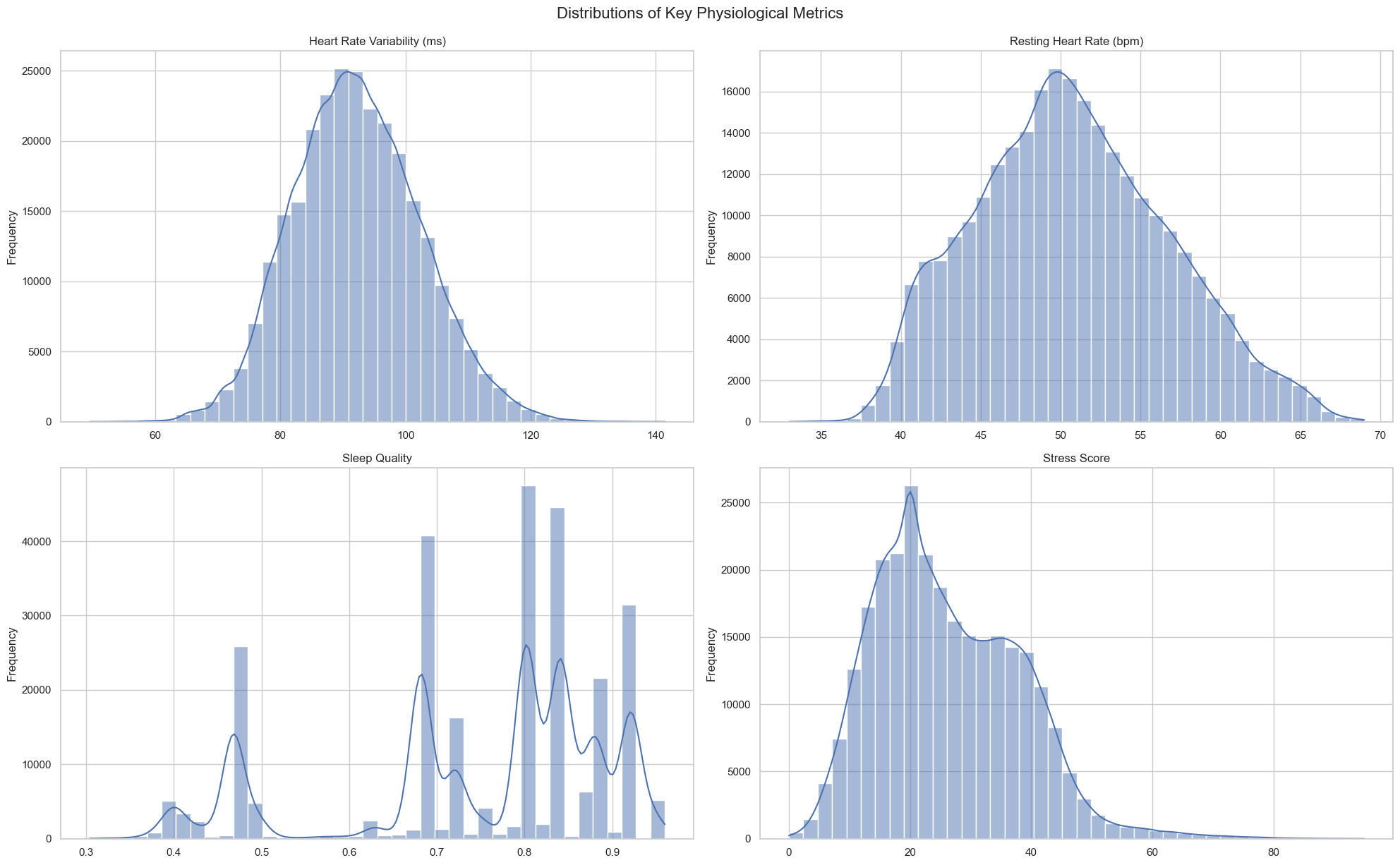}
\caption{Distribution of Key Physiological Metrics in Competitive Age-Group Triathlete Population. Top left: HRV; top-right: RHR; bottom-left: sleep quality; bottom-right: stress score.}
\label{fig: physiological-metrics}
\end{figure}

Fig.~\ref{fig: physiological-metrics} shows the empirical distributions of heart‑rate variability (HRV), resting heart rate (RHR), sleep quality, and stress score for the full population of \num{1000} simulated triathletes. HRV follows a "near" normal distribution with mean $92.3\text{\,ms}$ and standard deviation $10.2\text{\,ms}$, matching the range reported for endurance‑trained competitors by Plews et al. \cite{plews2013training} (80–140 ms). RHR is right‑skewed ($\gamma_{1}=0.24$) with a mean of $49\text{–}50\text{\,bpm}$, in which \SI{92.5}{\percent} of all values lie below the clinical bradycardia (condition of slow heart rate) threshold of \SI{60}{bpm}. Sleep quality exhibits three distinct modes at 0.50, 0.70, and 0.82 on a $[0,1]$ scale, reproducing the recovery categories (poor, adequate, optimal) published in the 2024 Garmin Connect report \cite{garmin2024data}.  The stress score is bimodal with peaks at 20 and 35 (Garmin scale 0–100) and mean $26.2\pm11.9$, close to the 2024 population average of~30.

\begin{figure}[h!]
\centering
\includegraphics[width=1\columnwidth]{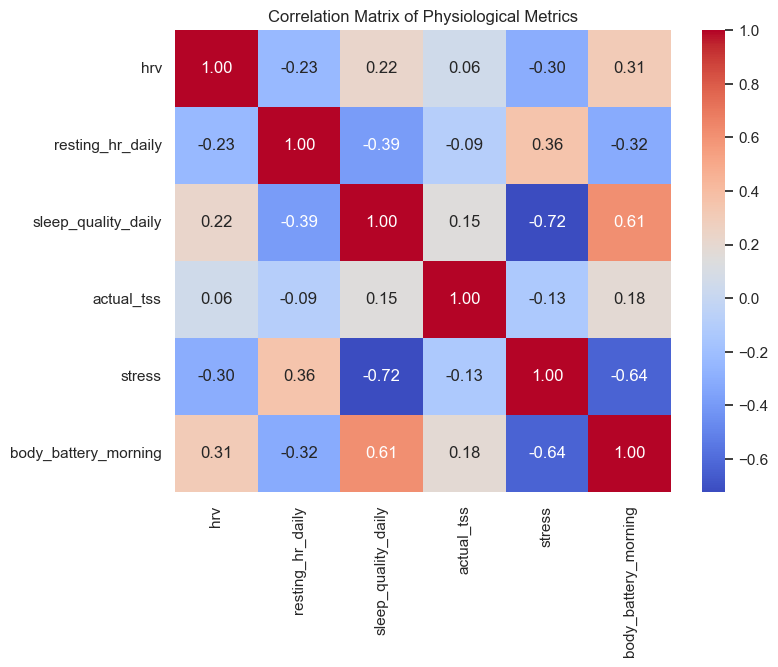}
\caption{Correlation Matrix of Physiological Metrics}
\label{fig: correlation-matrix}
\end{figure}

Inter‑metric relationships are summarised in Fig.~\ref{fig: correlation-matrix}.  HRV correlates negatively with RHR ($r=-0.23$), while sleep quality correlates negatively with stress ($r=-0.72$) and positively with Garmin’s body‑battery index ($r=0.61$).  These magnitudes are consistent with large cohort studies of autonomic balance during heavy training. \textit{Training Stress Score (TSS)} shows only weak positive associations with recovery indicators, indicating that fitter athletes sustain higher volumes without disrupting autonomic homeostasis.

\subsection{Pre-injury Realism (EQ2)}

Injuries in endurance sports rarely occur without warning. Rather, they typically emerge following a cascade of subtle physiological changes that can manifest days or even weeks before the injury becomes clinically apparent. The presence of these pre-injury patterns in the synthetic dataset, is crucial for developing effective early warning systems that could be used in real-world applications to allow athletes and coaches to intervene before tissue damage occurs. 

\begin{figure}[h!]
\centering
\includegraphics[width=1\columnwidth]{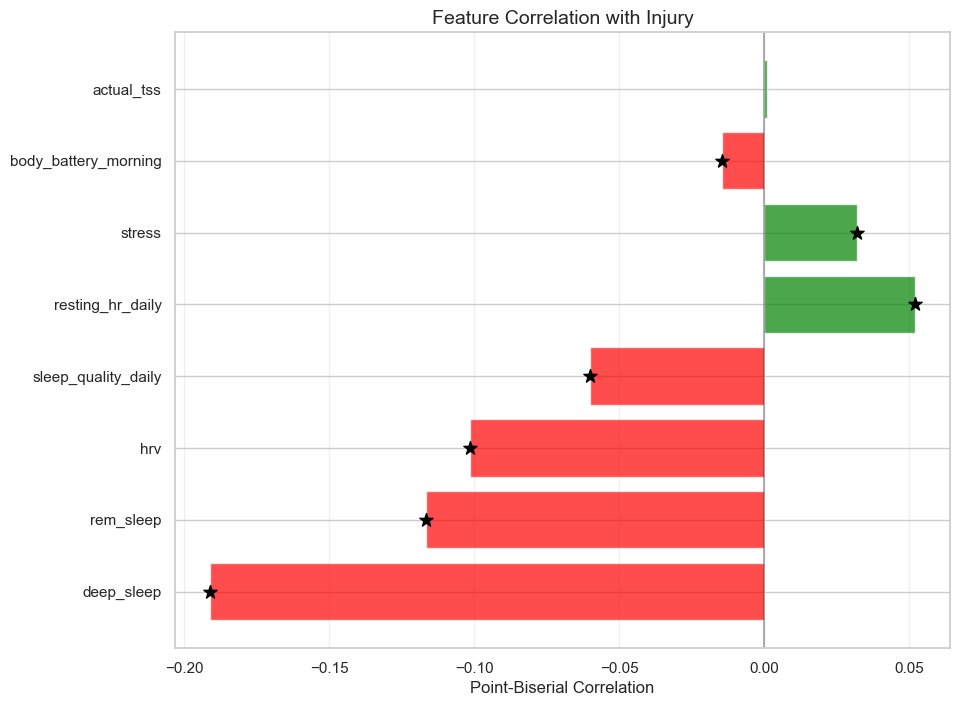}
\caption{Correlation of various Features with Injury Likelihood}
\label{fig: injury-correlation}
\end{figure}

\subsubsection{Feature Correlation with Injury} Fig.~\ref{fig: injury-correlation} presents point‑biserial coefficients between each metric and the binary injury flag. Recovery‑related variables dominate: deep‑sleep duration ($r\approx-0.19$) emerges as the \textbf{strongest} protective factor, followed by REM sleep, HRV, and composite sleep quality. Load‑related markers such as RHR ($r\approx0.06$) and perceived stress ($r\approx0.04$) display weaker but directionally consistent associations. These findings are consistent with research by Copenhaver and Diamond \cite{copenhaver2017value}, who found that sleep disturbance was significantly associated with injury risk in athletes.




In contrast, metrics associated with increased physiological load show positive correlations with injury risk. Resting heart rate demonstrates the strongest positive correlation (approximately 0.06), followed by stress levels (approximately 0.04). These positive correlations support work by Hellard et al. \cite{hellard2015training}, who found elevated resting heart rate to be a significant predictor of illness in swimmers, and Mann et al. \cite{mann2016effect}, who established relationships between psychological stress and injury in athletes.

\subsubsection{Temporal Progression of Warning Signs}

\begin{figure}[h!]
\centering
\includegraphics[width=1\columnwidth]{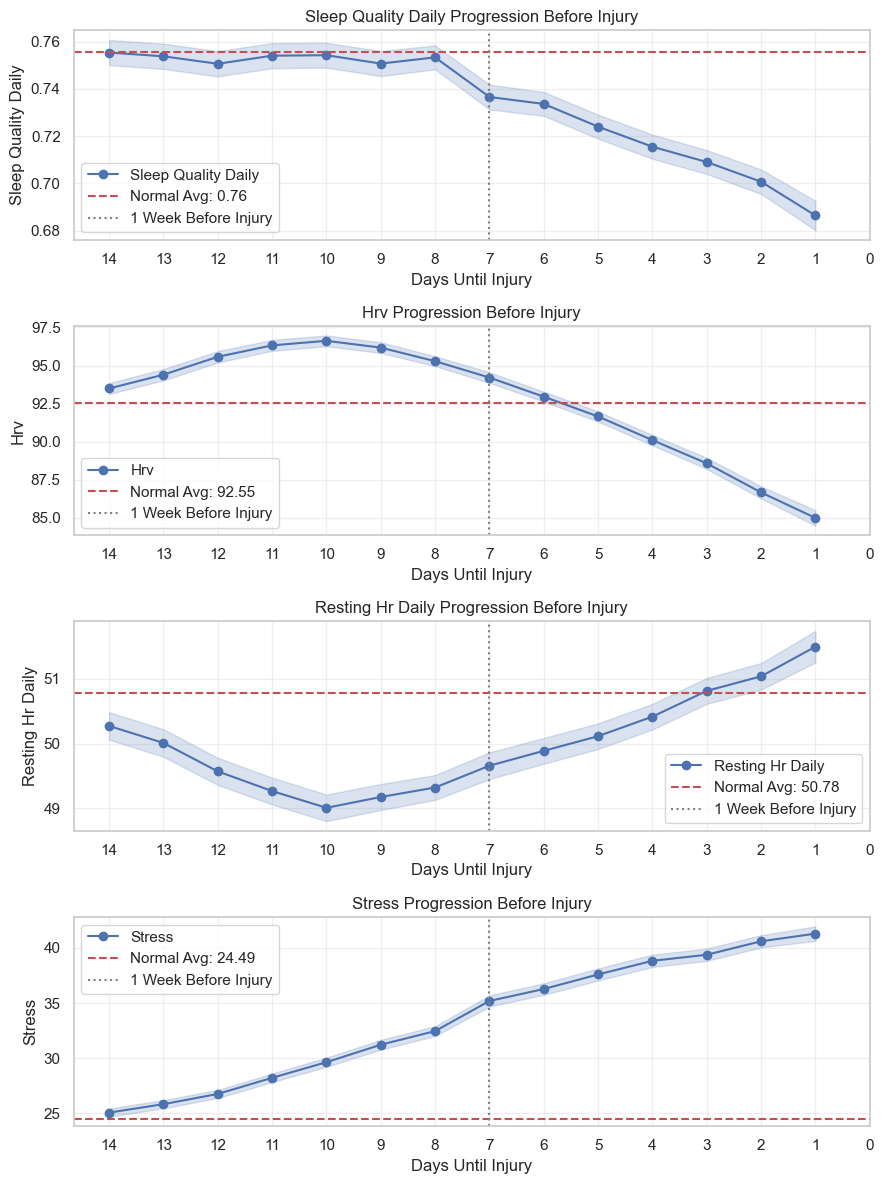}
\caption{Time Series Analysis of Recovery Metrics Before Injury}
\label{fig: pre-injury-pattterns}
\end{figure}

Figure \ref{fig: pre-injury-pattterns} illustrates the temporal progression of key recovery metrics, in the two-week period preceding injury occurrences across the synthetic dataset. 


Around seven days before injury (marked in the Figure \ref{fig: pre-injury-pattterns}), distinct physiological changes emerge, indicating a critical transition in recovery status. HRV exhibits a biphasic pattern with supranormal levels between days 14–8, followed by a steady decline toward injury, consistent with Plews et al.’s \cite{plews2013training} findings on maladaptive responses. Resting heart rate (RHR) inversely mirrors this trend, initially below average then rising after day 7 to exceed normal levels near injury onset. Sleep quality remains stable until day 7, after which it declines steadily alongside a continuous rise in stress levels, accelerating in the final week. This concurrent sleep deterioration and stress increase align with Hausswirth et al.’s \cite{hausswirth2014evidence} observations linking functional overreaching to elevated stress and impaired sleep.

\subsection{Predictive Benefit (EQ3)}

This goal should ensure that a realistic environment is created that reflects competitive age-group triathletes often relying on coaches that design their training following scientific principles in order to minimize injury risk and maximize performance gains. 

\subsubsection{Periodization Analysis} \label{periodization}
\begin{figure}[t!]
\centering
\includegraphics[width=1\columnwidth]{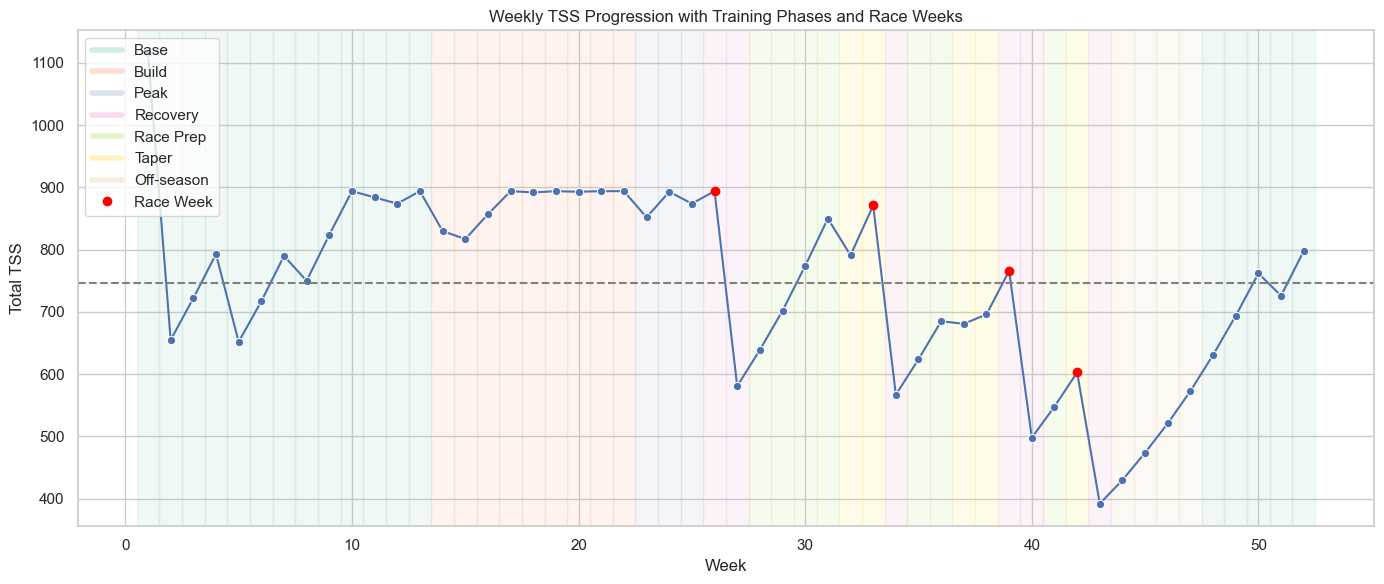}
\caption{Weekly TSS Progression with Training Phases}
\label{fig: tss-progression}
\end{figure}

The periodization analysis verifies controlled weekly training load progressions designed to optimize performance gains while minimizing injury risk. Figure \ref{fig: tss-progression} illustrates the planned weekly load alongside training phases and race weeks in the synthetic training plan.

Peak training loads reach about 900 TSS during build and peak phases, with sharp drops below 400 TSS in recovery and taper phases. Weekly increases rarely exceed the 10\% threshold, peaking at 10.256\% in week 30 during race preparation.

\subsubsection{Adherence to Load Management Principles}

Workload management is a critical aspect of injury prevention in endurance sports training. The \textit{Acute Workload Ratio (ACWR)} provides valuable insights into how current training load (acute) compares to established fitness (chronic), helping to identify periods of potential injury risk \cite{gabbett2016training}.




\begin{figure}[h!]
\centering
\includegraphics[width=1\columnwidth]{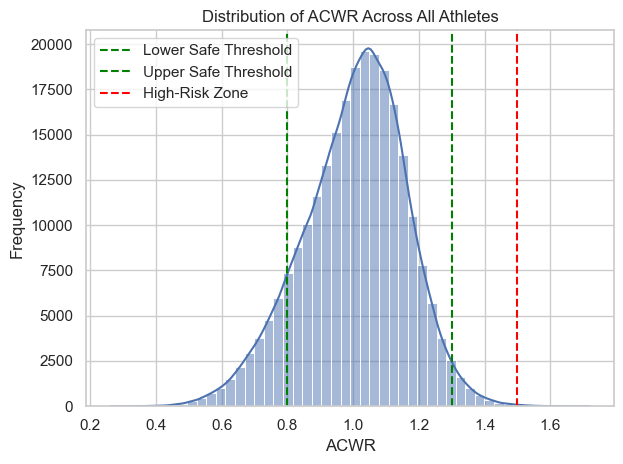}
\caption{Distribution of Acute Workload Ratio (ACWR) Across All Athletes}
\label{fig: acwr-distribution}
\end{figure}

Figure \ref{fig: acwr-distribution} reveals that across all simulated athletes, ACWR values follow a normal distribution centered around 1.0, with the vast majority falling within the widely accepted "safe zone" of 0.8 to 1.3 (indicated by the green vertical lines) \cite{gabbett2016training}. The red vertical line at 1.5 represents the high-risk zone threshold \cite{gabbett2016training}, which contains only a negligible portion of the distribution. This demonstrates that the synthetic training plans predominantly operate within established safe parameters for workload progression.

\subsubsection{Model Performance Evaluation}

Table \ref{tab:model_performance} presents the performance metrics for all three models in both data splitting strategies.

\begin{table}[!h]
\centering
\caption{Predictive performance (higher is better).}
\label{tab:model_performance}
\small
    \begin{tabular}{lcccc}
    \toprule
    \multirow{2}{*}{Model} & \multicolumn{2}{c}{Athlete‑based} & \multicolumn{2}{c}{Time‑based} \\
    \cmidrule(r){2-3}\cmidrule(l){4-5}
     & AUC & AP & AUC & AP \\
    \midrule
    LASSO (real)        & 0.823 & 0.664 & 0.826 & 0.671 \\
    LASSO (+synthetic)  & 0.849 & 0.710 & 0.855 & 0.725 \\
    RF (real)           & 0.830 & 0.653 & 0.833 & 0.661 \\
    RF (+synthetic)     & 0.853 & 0.700 & 0.855 & 0.718 \\
    XGB (real)          & 0.836 & 0.668 & 0.841 & 0.683 \\
    XGB (+synthetic)    & \textbf{0.852} & \textbf{0.709} & \textbf{0.858} & \textbf{0.726} \\
    \bottomrule
    \end{tabular}
\vspace{-0.3em}
\end{table}

XGBoost achieves the best overall performance (AUC = 0.858, AP = 0.726) across both splitting strategies, though differences among models are small (<1\% AUC variation). The comparable results from diverse algorithms—linear (LASSO), ensemble (Random Forest), and gradient boosting (XGBoost)—indicate that the synthetic dataset encodes clear, consistent injury patterns detectable by various approaches. This suggests a strong, well-structured predictive signal, validating the synthetic data’s quality. The high average precision (0.700–0.726) is particularly promising given the class imbalance, demonstrating the models’ ability to identify injury risk while controlling false positives, a key factor for real-world athlete monitoring.

\subsection{ML Feature Importance Analysis}

\begin{figure}[htbp]
    \centering

    \begin{subfigure}{\columnwidth}
        \includegraphics[width=\linewidth, trim={0 0 0 2cm},clip]{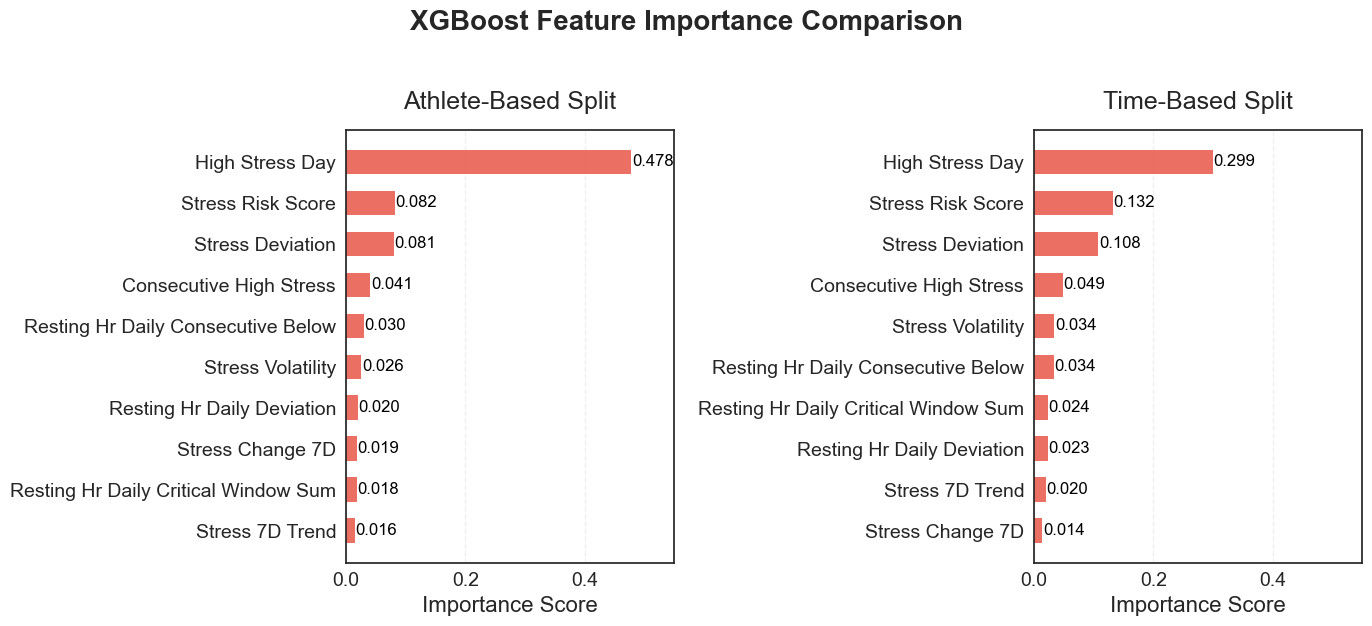}
        \label{fig:xgb_importance}
    \end{subfigure}
    
    \vspace{-2mm}
    
    \begin{subfigure}{\columnwidth}
        \includegraphics[width=\linewidth, trim={0 0 0 2cm},clip]{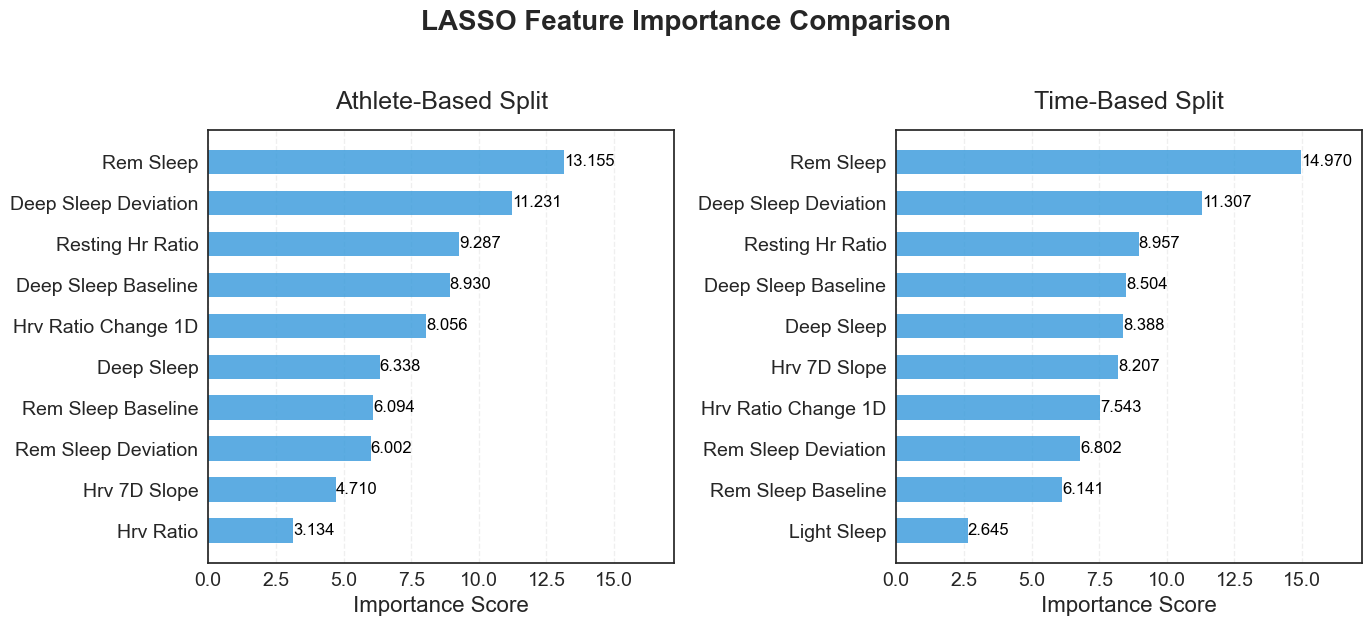}
        \label{fig:lasso_importance}
    \end{subfigure}
    
    \vspace{-2mm}

    \begin{subfigure}{\columnwidth}
        \includegraphics[width=\linewidth, trim={0 0 0 2cm},clip]{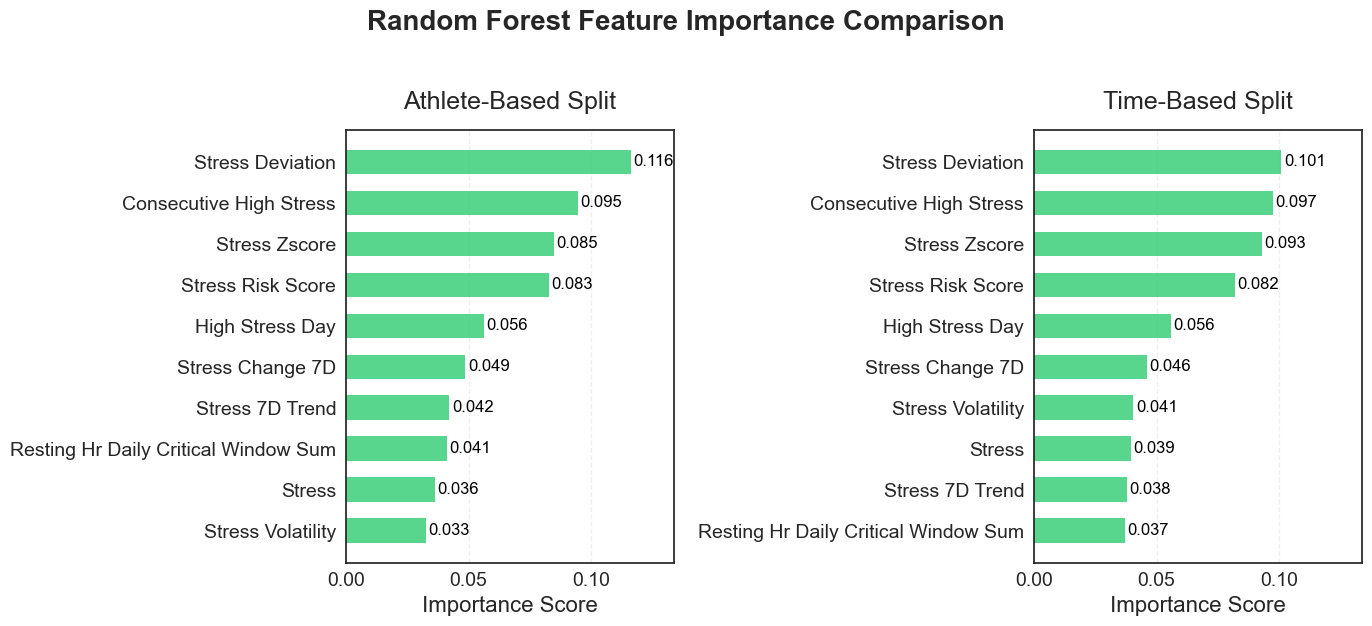}
        \label{fig:rf_importance}
    \end{subfigure}
    
    \caption{Feature importance by model: (a) XGBoost (top), (b) LASSO, (c) Random Forest (bottom).}
    \label{fig:feature-importance}
\end{figure}

The comparative analysis of feature importance reveals distinct yet complementary patterns in identifying key predictors of injury risk (cf. Fig. \ref{fig:feature-importance} - higher is more relevant).


\textit{Recovery-related metrics} consistently ranked as the most important predictors of injury risk, while training load indicators were absent. LASSO emphasized linear associations, with \textit{REM sleep} and \textit{deep sleep} deviation showing the highest weights in both splits, validating previous correlation analyzes. In contrast, Random Forest highlighted cumulative stress indicators such as \textit{stress deviation} and \textit{consecutive high stress days}, demonstrating its strength in capturing non-linear and interaction effects. XGBoost prioritized acute stress events particularly \textit{single high-stress days}, which varied in importance between splits, indicating that stress manifestations may differ across individuals. 


\subsection{Pathway to Real-World Deployment and Limitations} \label{transition-to-real-world}

We introduce a bridge for overcoming data limitations in triathlon, providing plausible synthetic data to progressively integrate real-world data (post-deployment). Despite efforts to ensure physiological plausibility, systematic differences between simulated data patterns and actual wearable device measurements could lead to degraded model performance when applied to real athletes. These differences can happen due to device-specific measurement characteristics, environmental factors not accounted for in the simulation, or subtle physiological relationships that have been oversimplified.

\section{Summary and Future Work}
\label{conclusion}

We presented a novel framework for injury prediction in triathletes, considering lifestyle factors. Unlike traditional models focused primarily on training loads, our holistic approach integrates stress, sleep, and recovery metrics, improving predictive personalization and generalizability. We defined key components of the synthetic data generation process, such as athlete profiling, periodized training modeling, and injury simulation, resulting in plausible athlete datasets. Future work will refine physiological realism, validate the synthetic framework against real-world data, and further assess machine learning models including LASSO, Random Forest, and XGBoost. All code, parameter files, and datasets are released \cite{rossiGitHub} to enable full replication of the results. 

\bibliographystyle{IEEEtran}
\balance
\bibliography{references.bib}

\end{document}